# Image Fusion and Re-Modified SPIHT for Fused Image

**S. Chitra, J. B. Bhattacharjee, B. Thilakavathi**
**Department of ECE, Rajalakshmi Engineering College, Chennai, affiliated to Anna University Chennai, India**

**ABSTRACT.** This paper presents the Discrete Wavelet based fusion techniques for combining perceptually important image features. SPIHT (Set Partitioning in Hierarchical Trees) algorithm is an efficient method for lossy and lossless coding of fused image. This paper presents some modifications on the SPIHT algorithm. It is based on the idea of insignificant correlation of wavelet coefficient among the medium and high frequency sub bands. In RE-MSPIHT algorithm, wavelet coefficients are scaled prior to SPIHT coding based on the sub band importance, with the goal of minimizing the MSE.

## INTRODUCTION

With the development of new imaging sensors arises the need of a meaningful combination of all employed imaging sources. The actual fusion process can takes place at different levels of information representation such as signal, pixel, feature and symbolic level. This site focuses on the so-called wavelet transform image fusion process, where a composite image has to be built of several input images. The most common form of transform image fusion is real valued wavelet transform fusion technique. All the input images are transformed and combined in the transform domain before an inverse transform results in the resultant fused image. The combination of the transformed image is achieved using a defined fusion rule. This rule can be as simple as choosing to retain the largest coefficient or more complicated windowed coefficient .The use of the real valued wavelet transform for image fusion has given good results especially when compared to naive pixel based and other transforms methods.

In recent years, many wavelet transform based image compression techniques with low reconstruction error and fine visual quality have been proposed. The effectiveness of the SPIHT (Set Partitioning In Hierarchical Trees) algorithm originates from the efficient subset partitioning and the compact form of the significance information. The SPIHT algorithm defines spatial orientation trees, sets of coordinates and the recursive set partitioning rules .The algorithm is composed of two passes: a sorting pass and a refinement pass. It is implemented by alternately scanning three ordered lists, called list of insignificant sets (LIS), list of insignificant pixels (LIP) and list of significant pixels (LSP), among which LIS and LIP represent the individual and sets of coordinates, respectively, whose wavelet coefficients are less than a threshold defined. During the sorting pass the significances of LIP and LIS are tested, followed by removal and set splitting operations to maintain the insignificance property of the lists. The LSP contains the coordinates of the significant pixels that are scanned in the refinement pass.

The weighted wavelet coefficients, which have a larger dynamic range than the original ones, make the SPIHT algorithm inefficient, because this results in more scans of the wavelet coefficients. Consequently, a Re-modified SPIHT is needed to overcome this problem. First, we note that the wavelet coefficients with 0 importance weights are blocked out. As a prior knowledge, we need not waste the bit budget on those wavelet coefficients. Therefore, after each sorting pass, we delete the corresponding coordinates from LIP and the sets whose components all have 0 importance weights from LIS. This is done at both the encoder and the decoder. The Goal of RE-MSPIHT is to minimize the MSE & improve the compression ratio.

## I. DISCRETE WAVELET TRANSFORM FUSION

The basic idea of all multiresolution fusion schemes is motivated by the human visual system being primarily sensitive to local contrast changes, e.g. the edges or corners. In the case of wavelet transform fusion all respective wavelet coefficients from the input images are combined using the fusion rule since wavelet coefficients having large absolute values contain the information about the salient features of the images such as edges and lines. The most common form of transform image fusion is wavelet transform fusion [Ber03, HBC05]. In common with all transform domain fusion techniques the transformed images are combined in the transform domain using a defined fusion rule then transformed back to the spatial domain to give the resulting fused image.

Wavelet transform fusion is more formally defined by considering the wavelet transforms ω of the two registered images $I_1(x, y)$ and $I_2(x, y)$ together with the fusion rule φ. Then the inverse wavelet transform $ω^{-1}$ is computed, and the fused image $I(x, y)$ is reconstructed:

$$I(x, y) = ω^{-1}(φ(ω(I_1(x, y)), ω(I_2(x, y))))$$

This process is depicted in figure 1.

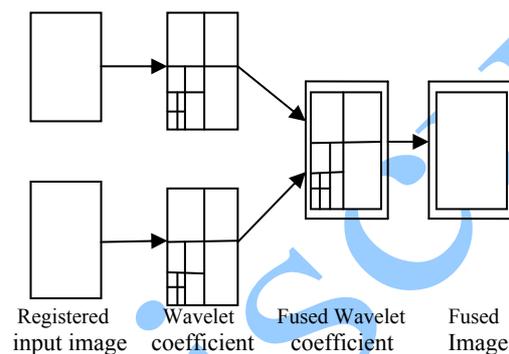

Registered input image    Wavelet coefficient    Fused Wavelet coefficient    Fused Image

**Figure 1. Fusion of the wavelet transforms of two images**

## A. Implemented Fusion Rules

Three fusion rule schemes [AZD07] were implemented using discrete wavelet transform based image fusion:

- Maximum selection (MS) scheme: This simple scheme just picks the coefficient in each Sub band with the largest magnitude;
- Principal component analysis (PCA): This scheme uses a normalized correlation between the two image sub bands over small local area. The resultant coefficient for reconstruction is calculated from this measure via a Eigen value and Eigen vector of two image coefficients
- Minimum selection (MS) scheme: This scheme just picks the coefficient in each sub band with the minimum magnitude.
- Fusion by averaging - for each band of decomposition and for each channel the wavelet coefficients of the two images are averaged

## B. Image Fusion Algorithm
**Step 1**
- Read first image in variable M1 and find its size (rows z l, columns: SI).

- Read second image in variable M2 and find its size (rows z2, columns: s2).
- Compare rows and columns of both input images. If the two images are not of the same size, select the portion, which are of same size.

**Step 2**

Apply wavelet decomposition and form spatial decomposition Trees. The one dimensional wavelet transform can be applied to the columns of the already horizontal transformed image as well. The image is decomposed into four quadrants with different interpretations.

LL: The upper left quadrant consists of all coefficients, which were filtered by the analysis low pass filter along the rows and then filtered along the corresponding columns with the analysis low pass filter again. This sub block is denoted by LL and represents the approximated version of the original at half the resolution.

HL/LH: The lower left and the upper right blocks were filtered along the rows and columns alternatively. The LH block contains vertical edges, mostly. In contrast, the HL blocks show horizontal edges very clearly.

HH: The lower right quadrant was derived analogously to the upper left quadrant but with the use of the analysis high pass filter which belongs to the given wavelet. We can interpret this block as the area, where we find edges of the original image in diagonal direction.

**Step 3**

The wavelet transforms $w$ of the two registered input images are combined utilizing some kind of fusion rule, given by $\varphi\ (\omega\ (I_1(x,\ y)), \omega\ (I_2(x,\ y)))$. For this fusion rules it has following methods,

Principal Component Analysis:

Compute, select & normalize Eigen value. If first Eigen value (1, 1)> Second Eigen value (2, 2), then the fusion matrix [a] becomes first Eigen vector/ Sum of the elements of first Eigen vector. Otherwise second Eigen vector/ Sum of the elements of second Eigen vector,

$$\text{Fused Image} = a\,[1]*M1 + a\,[2]*M2$$

Averaging Method:

$$\text{Fused Image} = (\omega\ (I_1(x,\ y)), \omega\ (I_2(x,\ y))\ /2$$

Minimum Method:

$$\text{Fused Image} = \text{MIN}\ \{\omega\ (I_1(x,\ y)), \omega\ (I_2(x,\ y))\}$$

Maximum Method:

$$\text{Fused Image} = \text{MAX}\{\omega(I_1(x, y)), \omega(I_2(x, y))\}$$

Compute The Entropy Of Fused Image:

$$\text{ENTROPY} = -\sum (P(x, y) * \log_2(P(x, y)))$$

where $P_i$ is the probability for each element
**Step 4**
  Apply Inverse transformation to give the fused image. However, the rules must be applied to the magnitude of the DWT transformed coefficients.

$$I(x, y) = \omega^{-1}(\varphi(\omega(I_1(x, y)), \omega(I_2(x, y)))).$$

## II. SPIHT ALGORITHM

SPIHT algorithm, introduced by A. Said and W.A. Pearlman, adopts a hierarchical quad tree data structure on wavelet-based image. Figure 2 indicates the parent-child relationship through the sub bands (quad tree). The original SPIHT [CC06] is briefly described as follows. The wavelet coefficient are encoded and transmitted in multiple passes in the SPIHT algorithm.
**Step 1**
  In each pass, only the wavelet coefficients that exceed threshold are encoded. The threshold $T(u)$ is computed according to the expression

$$T(u) = 2^{P-u} \ldots\ldots\ldots(1)$$

where u=0, 1, 2, 3… P denotes the pass number, and

$$P = \log_2 \max(c(i,j)) \ldots\ldots(2)$$

where $c(i,j)$ is the coefficient at position$(i,j)$ in the image.

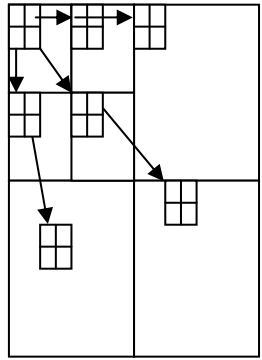

**Figure 2 Quad tree organization of wavelet coefficients in SPIHT algorithm.
The above figure shows the Parent-child Relationship**

It just sends the max value to the decoder, and the thresholds can be calculated by (1) and (2).
**Step 2:** Sorting pass
When u=n, n is integer. The pixel that satisfying $T(n) \leq |c(i,j)| < 2T(n)$ is identified as significant. Where c (i,j) is coefficient value. The pixel's position and sign bit must be encoded.
**Step 3:** Refinement pass
The pixels that satisfying
$c(I,j) \geq 2T(n)$ are refined by encoding the nth most significant bit(those that had their coordinates transmitted in previous sorting passing).
**Step 4**: Increment u by one and go to step 2.

## III. RE-MODIFIED SPIHT

The set partitioning in hierarchical trees (SPIHT), an efficient wavelet-based progressive image-compression scheme, is oriented to minimize the mean-squared error (MSE) between the original and decoded imagery. In this paper, first estimate the importance of each wavelet sub-band for distinguishing between different textures segmented by an HMT mixture model. Figure 3 shows the overall Re-modified SPIHT (RE-MSPIHT) coding scheme [CC06, Yen05]. Before the SPIHT coding, we weight the wavelet coefficients, with the goal of achieving improved image-classification results at low bit rates. A Re-modified SPIHT algorithm is proposed to improve the coding efficiency. The performances of the original

SPIHT and the Re-modified SPIHT algorithms are compared. Hidden Markov trees (HMT) in the wavelet domain capture the statistical dependence of wavelet coefficients well, providing a reliable segmentation of image textures. The segmentation is performed autonomously at the encoder, and the goal is to prioritize for compression those wavelet coefficients that play an important roles in this segmentation stage, despite the fact that these coefficients may be of small amplitude (and hence given low priority by conventional wavelet encoders). Set partitioning in hierarchical trees (SPIHT) is an effective embedded wavelet-based image-coding algorithm. It seeks to minimize the mean-squared error (MSE) at any bit rate by the progressive transmission of the partially ordered bit planes and the effective exploration of the self-similarity across the wavelet trees. However, the MSE-based measure is not in general well correlated with the image-recognition quality, especially at low bit rates (small wavelet coefficients, which may be important for classification, are given low priority and therefore a coarse representation by conventional MSE-based encoders). To optimize the image visual quality, perceptually weighted quantization demonstrates a significant improvement in visual quality. By weighting the wavelet coefficients properly, we order the transmission bits not only by the magnitudes of the wavelet coefficients but also by their contributions to image recognition. An efficient algorithm is developed especially for the weighted wavelet coefficients, to improve the compression ratio.

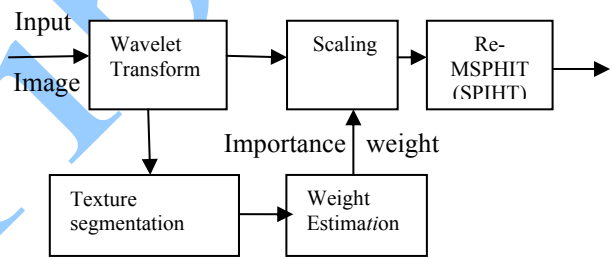

**Figure 3 Re-Modified SPIHT (MSPIHT) coding scheme**

### A. TEXTURE SEGMENTATION

Image segmentation is a fundamental low-level operation in image analysis for object identification. Clustering can be considered the most important unsupervised learning problem [Moo01], so it deals with finding a structure in a collection of unlabeled data shown in Figure 4. A cluster is therefore a collection of objects which are "similar" between them and are "dissimilar" to the objects belonging to other clusters.

Clustering algorithms may be classified as listed below:
- Exclusive Clustering
- Overlapping Clustering
- Hierarchical Clustering
- Probabilistic Clustering

The procedure follows a simple and easy way to classify a given data set through a certain number of clusters (assume k clusters) fixed a priori. The main idea is to define k centroids, one for each cluster. These centroids should be placed in a cunning way because of different location causes different result. So, the better choice is to place them as much as possible far away from each other. The next step is to take each point belonging to a given data set and associate it to the nearest centroid. When no point is pending, the first step is completed and an early groupage is done. This algorithm aims at minimizing an objective function, in this case a squared error function.

Algorithm [WZ05]:
- Ask user how many clusters
- Randomly guess cluster Center  locations
- Each data point finds out which Center it's closest to.
- . …and jumps there
- …Repeat until terminated.

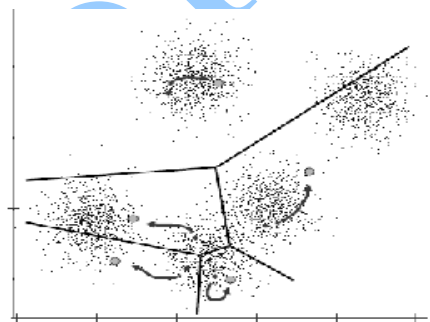
**Figure 4 Clustering scheme**

**HMT MODEL**

The HMT model [CNB98, SF00] is a tree-structured probabilistic graph that captures the statistical properties of the wavelet transforms of images. This technique has been successfully applied to the segmentation of natural texture images, documents etc. In a tree-structured HMT model, was proposed in the wavelet domain to achieve the statistical characterization of signals and images by capturing interscale dependencies of wavelet coefficients across scales. It is a two-step procedure: (1) Estimate parameters in different regions and find a coarse segmentation of the image; (2) Estimate the parameters again in each segmented region, then use updated parameters to perform supervised segmentation. In those methods, a local window which divides given image into small sub images is necessary to the parameter estimation, and sometimes the window size is changeable under multiresolution framework. The wavelet transform provides a natural multiscale structure of signal/image analysis, and wavelet domain HMT models can characterize image features in the wavelet domain for the unsupervised segmentation. If the image contains more than one texture, we use HMT mixtures to model the statistical characteristics.

The probabilistic model is

$$p(w/\emptyset) = \sum_{i=1}^{M} \alpha_i \, P(w/HMT_i)$$

where $\alpha_i \geq 0$, $\sum \alpha_i = 1$ are the mixing coefficients of the M textures, which can be interpreted as the prior probabilities and w are the wavelet coefficients of a wavelet tree. P (w/HMT$_i$) is a density function parameterized via the HMT for texture i. We estimate the probability that the t$^{th}$ wavelet tree is generated by texture j using Bayes'srule

$$P_j^k(t) = \frac{\alpha_j \, P(w_t/HMT_i)}{\sum_{i=1}^{M} \alpha_i P(w_t/HMT_i)}$$

and update the mixing coefficient as

$$\alpha_j^{(k+1)} = \frac{\sum_{t=1}^{N} P_j^k(t)}{N}$$

$t=1, 2\ldots N$.

The parameters of each HMT are updated using the sample probability $P_j^{(k)}$ (t),t=1,2,….,N. The samples that are associated with texture *j* with high likelihood make a greater contribution to the parameters of that texture component. We iteratively estimate the probabilities and update the model parameters until the model converges to a local optimal solution. The compression property of the wavelet transform states that the transform of a typical signal consists of a small number of large coefficients and a large number of small coefficients. More precisely, most wavelet coefficients have small values and, hence, contain very little signal information. A few wavelet coefficients have large values that represent significant signal information. This property leads to the following simple model for an individual wavelet coefficient. We model each coefficient as being in one of two states: "high," corresponding to a wavelet component containing significant contributions of signal energy, or "low," representing coefficients with little signal energy. If we associate with each state a probability density—say a high-variance, zero-mean density for the "high" state and a low-variance, zero-mean density for the "low" state—the result is a two-state mixture model for each wavelet coefficient. In most applications of mixture models, the value of the coefficient is observed, but the value of the state variable is not; we say that the value of is *hidden*. In Fig.5.each black node represents a continuous wavelet coefficient $w_i$. Each white node represents the mixture state variable $S_i$ for $w_i$. To match the intercoefficient dependencies, we link the hidden states. Connecting discrete nodes horizontally across time (dashed links) yields the hidden Markov chain model. Connecting discrete nodes vertically across scale (solid links) yields the HMT model.

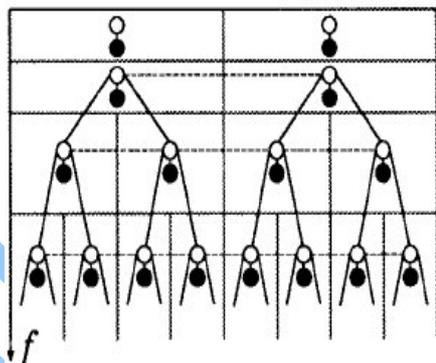

**Figure 5 Statistical models for the wavelet transform**

## B. IMPORTANCE WEIGHT ESTIMATION

Wavelet coefficients prioritized based on their amplitude strength may not optimize the recognition rate (here characterized by distinguishing textures) based on the decoded imagery, so we introduce a rescaling process using estimated importance weights [CC04]. The purpose of the importance weights is to help the encoder to order the output bit stream with consideration of the ultimate recognition task. The wavelet features that are highly correlated with the texture class labels are important to distinguish the textures. These features should have higher priorities and be transmitted earlier. However, the wavelet features that are more important to segmentation may have small amplitudes and small variances compared to the wavelet features of less importance to segmentation. The rescaling process alleviates this phenomenon.

**RE-MODIFIEDSPIHT ALGORITHM**:
I case:
- Compute threshold (T) from LL band.
- Compare 'T' with LH, HL, and HH band coefficients.
- Retaining the coefficients that are correlated with three bands in the same level.
- Apply SPIHT algorithm to these coefficients.

II case:
- Compute entropy of each cluster.
- Count the non-zero elements in each cluster.
- ENTROPY= -Sum(Pi*log2(Pi)) where Pi is the probability for each element
- Find the importance weight of each wavelet Coefficients
- Wavelet coefficients with 0 importance weights (less number of occurrences) are blocked out.
- Retaining the most important coefficients- i.e. the wavelet coefficients with high probability of occurrence is given to scaling process.
- The important coefficients are scaled to large relative amplitude by multiplying these coefficients with some scaling factor.
- The scaled coefficients are applied to SPIHT algorithm.

In the generative learning progress, we select the wavelet coefficients one by one to reduce the cost function defined as follows [VB02]:

$$e = \frac{1}{N}\sum_{i=1}^{N}(\lambda(Y_i - f(\hat{w}_i))^2 + \|w_i - \hat{w}_i\|^2)$$

The importance weight of the i[th] selected wavelet coefficients is defined as:

$$\beta_i = \begin{cases} -1 + \lambda.|\gamma_i| & i < m \\ 0 & otherwise \end{cases}$$

where λ is a Lagrangian multiplier and **w**ˆ is the quantized wavelet coefficients associated with the *i*-th wavelet tree.

## IV. SIMULATION RESULT

Image Fusion and modified SPIHT code are simulated using Matlab 6.5. The simulation output for both Fusion and RE-MSPIHT are as follows:

### A. Experiments and Results - Fusion Method

Meaningful comparison of [WZ05] is often dependant on the application. For some applications (e.g. medical image fusion) the aim of the fusion is to combine perceptually salient image elements such as edges and high contrast regions. Evaluation of fusion techniques for such applications can only be effectively based on a perceptual comparison. Figure 6 shows the experimental results of fusing multifocal images by different fusion rules For other applications (such as multifocal image combination) computational measures can also be used for fusion evaluation.

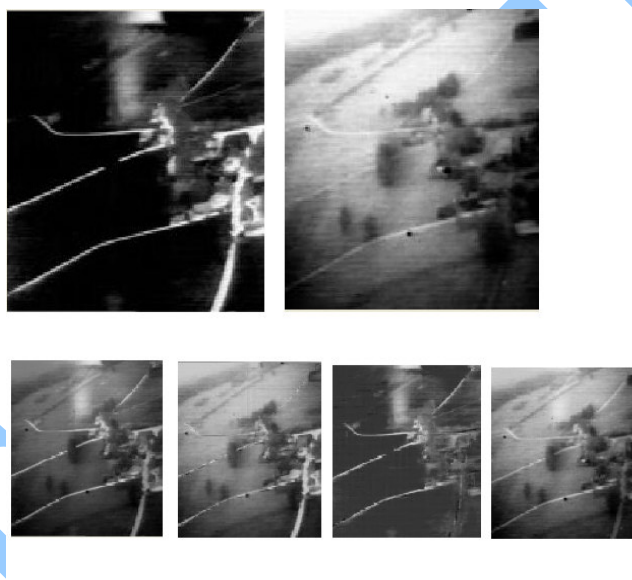

**Figure 6 Experimental results of different fusion rules. (a) Image I (b) Image 2 (c) Averaging (b) Maximum (c) Minimum (d) PCA method**

**Table 1: Entropy performance comparison of image fusion results for image of (256*256)**

| Fusion methods | Entropy |
|---|---|
| Averaging method | 10.88 |
| Maximum method | 11.414 |
| Minimum method | 11.516 |
| Principal component analysis | 14.262 |

154

## B. Experiments and Results - Re-Modified SPIHT

Between the wavelet transform and the SPIHT coding, we introduce a rescaling process, assigning finer quantization step sizes to the wavelet features with larger importance weights. After weighting the wavelet coefficients shown in Figure 7 as designed in the previous section, the SPIHT algorithm may be less efficient for implementation of the spatial orientation zero tree structure (SPIHT was originally designed for a MSE cost function alone; the new weighting of importance to potentially small wavelet coefficients may undermine assumptions in designing SPIHT).We modify the SPIHT algorithm to improve the coding efficiency, PSNR, Compression ratio and reduce the MSE.

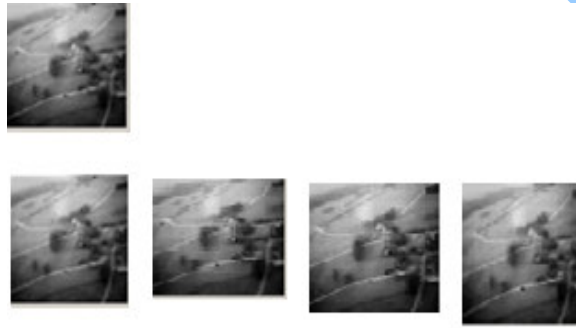

**Figure 7 Image quality a) Input image b) JPEG c) JPEG 2000 d) SPIHT e)RE-MSPIHT**

**Table 2: Performance comparison of image compression**

| Parameter (db) | JPEG | JPEG 2000 | SPIHT | RE-MSPIHT |
|---|---|---|---|---|
| PSNR | 38.84 | 33.79 | 41.93 | 62.04 |
| MSE | 8.54 | 15.16 | 4.19 | 0.04 |
| CR | 1.45 | 1.74 | 2.15 | 3.13 |

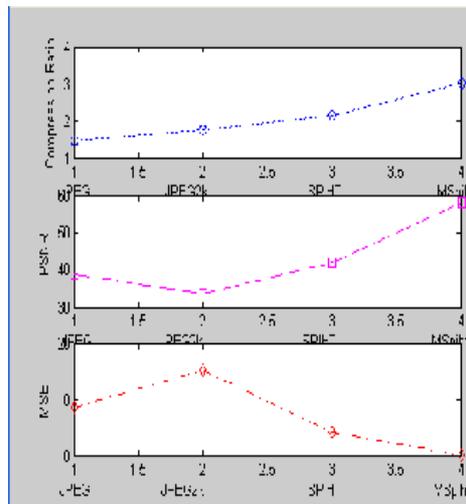

**Figure 8 Comparison chart**

## CONCLUSION

In this paper we have presented the Wavelet based Image Fusion and Re-Modified SPIHT algorithm. With the ultimate goal of image classification, we designed a scheme for properly pruning and weighting the wavelet coefficients before wavelet coding. We proposed a Re-modified SPIHT algorithm, using the importance weight information, to save the bit budget and reduce the MSE. Image Fusion can be applied to Navigation aid, merging out of focus images, remote sensing and medical images. RE-MSPIHT algorithm can be applied not only to still images; it can be applied for Video conferencing and mobile multimedia data services.

**Future enhancement**

In future this work can be extended to streaming video, satellite and 3D images. The wavelet transform used in the system developed here employed HAAR wavelet as a basis. The algorithm can be analyzed for different wavelet families. The work done here is for a gray scale image. It can be extended to color images. In real time it can be implemented in FPGA.